\documentclass[10pt,twocolumn,letterpaper]{article}

\usepackage[pagenumbers]{cvpr} 
\usepackage[utf8]{inputenc}
\usepackage{amsmath}
\usepackage{amsfonts}

\definecolor{cvprblue}{rgb}{0.21,0.49,0.74}
\usepackage[pagebackref,breaklinks,colorlinks,allcolors=cvprblue]{hyperref}

\def\paperID{8792} 
\def\confName{CVPR}
\def\confYear{2025}

\title{OMENN: One Matrix to Explain Neural Networks}

\author{Adam Wróbel$^{1,2}$\\
\and
Mikołaj Janusz$^1$\\
\and
Bartosz Zieliński$^1$\\
\and
Dawid Rymarczyk$^{1,3}$\\
\and
$^1$Jagiellonian University, Faculty of Mathematics and Computer Science\\
$^2$Jagiellonian University, Doctoral School of Exact and Natural Sciences
$^3$Ardigen SA \\ 
}

\begin{document}
\maketitle
\begin{abstract}
Deep Learning (DL) models are often black boxes, making their decision-making processes difficult to interpret. This lack of transparency has driven advancements in eXplainable Artificial Intelligence (XAI), a field dedicated to clarifying the reasoning behind DL model predictions. Among these, attribution-based methods such as LRP and GradCAM are widely used, though they rely on approximations that can be imprecise.

To address these limitations, we introduce One Matrix to Explain Neural Networks (OMENN), a novel post-hoc method that represents a neural network as a single, interpretable matrix for each specific input. This matrix is constructed through a series of linear transformations that represent the processing of the input by each successive layer in the neural network. As a result, OMENN  provides locally precise, attribution-based explanations of the input across various modern models, including ViTs and CNNs. We present a theoretical analysis of OMENN based on dynamic linearity property and validate its effectiveness with extensive tests on two XAI benchmarks, demonstrating that OMENN is competitive with state-of-the-art methods.

\end{abstract}   

\section{Introduction}
\label{sec:intro}


Deep learning (DL) has revolutionized fields such as computer vision~\cite{voulodimos2018deep} and natural language processing~\cite{otter2020survey}, enabling models to match or even surpass human performance in tasks such as image recognition and machine translation. However, this success comes at a cost. Deep neural networks operate as black boxes~\cite{rudin2019stop}, making their decision-making processes difficult to explain.

\begin{figure}
    \centering
    \includegraphics[width=0.48\textwidth]{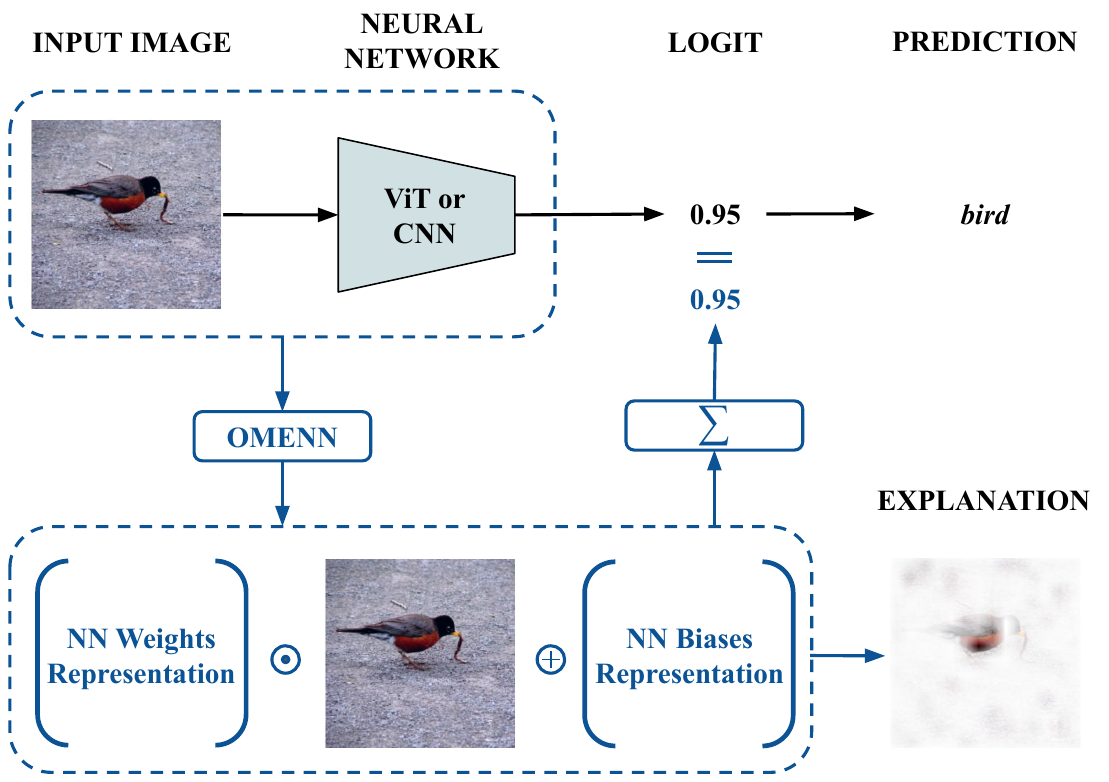}
    \caption{This figure illustrates OMENN, a novel XAI method that provides locally exact explanations by transforming the neural network's processing of an input into a linear operation comprising two matrices: one representing the model's weights and the other representing its biases. When combined with the input, these matrices yield a single, interpretable matrix that forms the explanation. Note that the sum of the values within this explanation matrix directly corresponds to the logit value of the class being explained.}
    \label{fig:teaser}
    \vspace{-1.5em}
\end{figure}

The black-box nature of deep neural networks allowed the development of eXplainable Artificial Intelligence (XAI)~\cite{samek2019towards}, a field focused on creating tools to reveal and interpret the reasoning processes of these models. Among XAI techniques, attribution-based methods, such as Layer-wise Relevance Propagation (LRP)~\cite{bach2015pixel} and GradCAM~\cite{chattopadhay2018grad}, are widely used by practitioners~\cite{abusitta2024survey}. These methods generate visual attribution maps that highlight regions in an image deemed significant for a given prediction.

However, attribution-based approaches often rely on indirect approximations~\cite{rudin2019stop}, such as gradients, to infer which parts of an input may be important for the model’s decision-making. By using gradients or other heuristics to deduce influential regions, they can introduce biases, resulting in explanations that are sometimes unreliable, as shown by various sanity checks~\cite{adebayo2018sanity,hedstrom2024fresh,tomsett2020sanity}.

To overcome this limitation, we introduce One Matrix to Explain Neural Networks (OMENN), a novel post-hoc method that precisely explains neural network decisions. OMENN represents a neural network for a specific input as a single, interpretable matrix. This is accomplished by reformulating the network as a series of linear transformations applied to a given data point, similar to the B-Cos model~\cite{bohle2022b}, but without requiring any modifications to the model architecture. Through this approach, we can decompose the network’s output into a linear transformation of the input, enabling explanations that are directly derived from the model’s weights and provide an exact representation of its decision process. This derivation is possible due to the property of dynamic linearity, which allows OMENN to represent a neural network as a single matrix for each specific input. Figure~\ref{fig:teaser} illustrates OMENN's intuition.

Although efforts have been made to express neural networks as a single matrix, existing methods face notable limitations. For instance, B-cos~\cite{bohle2024b} requires specialized architectures that exclude biases, while FullGrad~\cite{srinivas2019full} accommodates biases but is restricted to networks with only ReLU or LeakyReLU activation functions. In contrast, OMENN addresses this, supporting a wider range of architectures and activation functions.

We provide theoretical justification for OMENN and conduct extensive testing using the FunnyBirds~\cite{hesse2023funnybirds} and Quantus~\cite{hedstrom2023quantus} benchmarks, as well as a toy example, showing that our approach achieves results competitive to the state-of-the-art methods. We also present qualitative comparisons to showcase the distinctions between our method and existing approaches.

\begin{figure*}
    \includegraphics[width=0.95\textwidth]{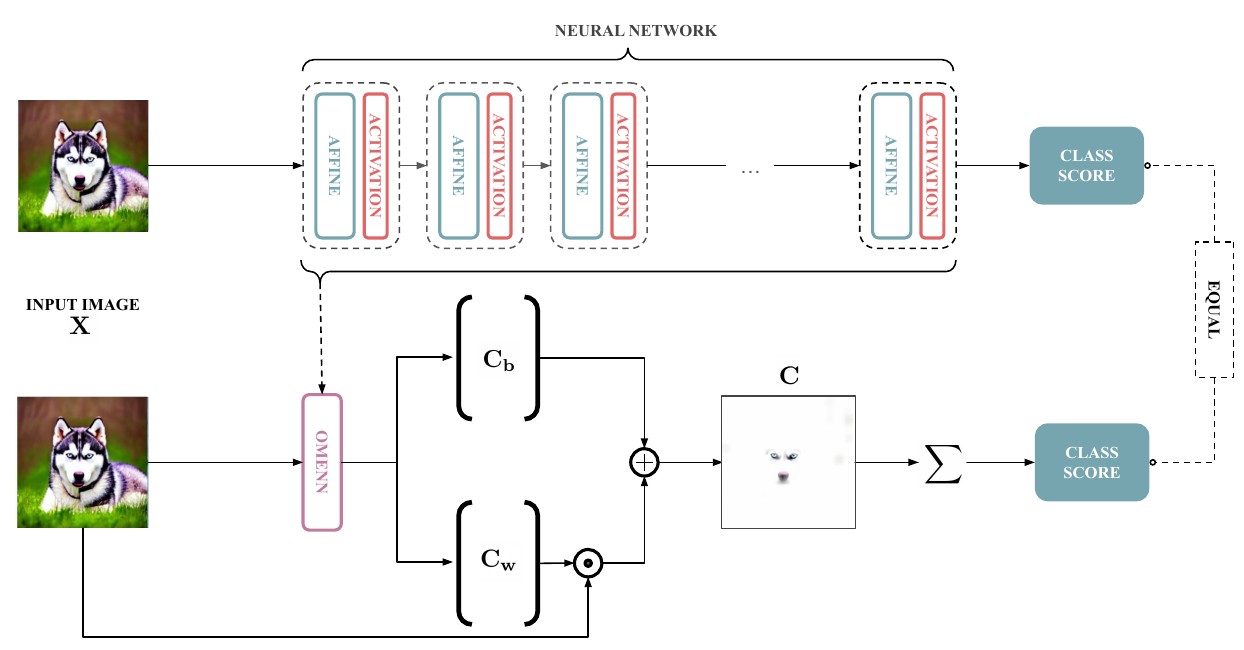}
    \caption{
OMENN method provides an explanation in the form of a single matrix, \(\bold{C}\). For this purpose, it represents a neural network as a single, input-dependent affine transformation using dynamic linearity. This transformation includes two components: a bias representation denoted as \(\bold{C_b}\) and a weight representation \(\bold{C_w}\). These components are applied to the input as an affine transform to form the overall explanation matrix \(\bold{C} = \bold{C_w}\odot\bold{X} + \bold{C_b}\). Notably, the sum of all elements in \(\bold{C}\) matches the score produced by the original neural network, ensuring the explanation is faithful to the network’s output.
    }
    \label{fig:overview}
\end{figure*}

Our contributions may be summarized as follows:
\begin{itemize}
    \item We present a method to represent a broad class of modern deep neural networks (e.g., ViTs and CNNs) as a single matrix by leveraging dynamic linearity.
    \item We introduce OMENN, a novel attribution-based XAI technique that uses this single-matrix representation to reveal the contribution of individual input image pixels to the final prediction score.
    \item We provide both theoretical and experimental evidence demonstrating OMENN’s competitive performance.
\end{itemize}

\section{Related Works}
\label{sec:rw}

The primary aim of eXplainable Artificial Intelligence (XAI) is to clarify the decision-making processes of deep neural networks~\cite{abusitta2024survey}. However, XAI methods extend beyond simply explaining model predictions to users; they also support broader objectives, such as enhancing continual learning~\cite{dhar2019learning} and improving distillation techniques~\cite{parchami2024good}.

XAI approaches generally fall into two categories: ante-hoc and post-hoc methods~\cite{rudin2019stop}. Ante-hoc methods focus on developing models with built-in transparency, offering explanations as part of the prediction process. These methods include techniques that generate contribution maps of the input, such as B-Cos~\cite{bohle2022b,bohle2024b} and Self-Explainable Neural Networks (SENN)~\cite{alvarez2018towards}, as well as concept-based approaches like Concept Bottleneck Models~\cite{koh2020concept}, which rely on concept supervision, and prototypical part-based models~\cite{chen2019looks,nauta2023pip,pach2024lucidppn,rymarczyk2021protopshare,rymarczyk2022interpretable} that identify key elements within a training dataset to justify decisions.

Post-hoc methods, on the other hand, are applied to trained black-box models, aiming to interpret predictions without altering the model architecture~\cite{abusitta2024survey}. Examples include Concept Activation Vectors (CAV)~\cite{ghorbani2019towards,kim2018interpretability,kowal2024visual}, which use predefined human concepts to explain their contribution to the model's output, perturbation-based methods~\cite{fel2023don,fong2017interpretable,li2021towards} that modify the input to reveal influential features, and local approximation techniques~\cite{ribeiro2016should} that identify linear decision boundaries in the model’s neighborhood. Other post-hoc methods involve counterfactual explanations~\cite{augustin2022diffusion,goyal2019counterfactual,jacob2022steex,jeanneret2023adversarial}, which show how input modifications would alter the model’s prediction, and attribution maps~\cite{bach2015pixel,chattopadhay2018grad,chefer2021transformer,gautam2023looks,montavon2019layer,selvaraju2017grad,srinivas2019full,sundararajan2017axiomatic}, which highlight regions of the input that are significant for a particular prediction.


\vspace{-1em}

\paragraph{Attribution-based XAI.}

Attribution-based explainers are designed to identify which parts of the input data were most influential in a model's decision. Key methods include Layerwise Relevance Propagation (LRP)~\cite{bach2015pixel}, which uses Taylor expansion to calculate neuron contributions, and gradient-based techniques such as GradCAM~\cite{selvaraju2017grad}, Integrated Gradients~\cite{sundararajan2017axiomatic}, and Gradient NN Representation~\cite{srinivas2019full}, which approximate important input regions using gradients. Another approach, Shapley values~\cite{zheng2022shap}, draws on cooperative game theory to assign contributions to input features. However, these methods have demonstrated limitations in reliability~\cite{adebayo2018sanity,binder2023shortcomings,hedstrom2024fresh}, as revealed by multiple sanity checks highlighting inconsistencies and inaccuracies.

To address these challenges, inherently interpretable methods with attribution-like explanations have been developed. For example, Self-Explainable Neural Networks (SENN)~\cite{alvarez2018towards} extract interpretable concepts that uphold fidelity, diversity, and grounding, combining them linearly to make decisions. This approach was later refined to QSENN~\cite{norrenbrock2024q}, which classifies each feature's influence as positive, negative, or neutral for each class, enhancing interpretability. Another alternative, the B-Cos model~\cite{bohle2022b,bohle2024b}, uses cosine-based non-linearity to align input features with the model output, improving transparency. However, both QSENN and B-Cos require architectural modifications and retraining from scratch, making them incompatible with pre-trained models.


\section{OMENN}



Our goal is to demonstrate that each layer in modern neural network architectures, such as Vision Transformer~\cite{dosovitskiy2021imageworth16x16words} or ConvNeXt~\cite{liu2022convnet}, can be expressed as an affine or input-dependent affine transformation. This property carries significant practical implications, as it enables the representation of the entire neural network as a single input-dependent affine transformation.

The OMENN algorithm (see Figure \ref{fig:overview}) decomposes the neural network to two input-dependent interpretable matrices: the weight representation matrix \(\bold{C_w}\) and the bias representation matrix \(\bold{C_b}\). To formalize this concept, let us define a sequential input \(\bold{X} \in \mathbb{R}^{t_{in} \times d_{in}}\) (where for image data \(t_{in}\) represents spatial dimensions \(hw\), and \(d_{in}\) denotes the channel dimension \(c\)). The algorithm takes an already trained neural network \(f: \mathbb{R}^{t_{in} \times d_{in}} \rightarrow \mathbb{R}\), and reformulates it into a single, input-dependent affine transformation applied to \(\bold{X}\). Specifically, it generates a weight representation matrix \(\bold{C_w} \in \mathbb{R}^{t_{in}\times d_{in}}\) and a bias representation matrix \(\bold{C_b} \in \mathbb{R}^{t_{in} \times 1}\), and combines them with the input to a single explanation matrix \(\bold{C}\), satisfying the following relationship:

 \begin{equation}
     \bold{C} = \sum_{d}(\bold{C_w}\odot\bold{X})[:,d] + \bold{C_b}
 \end{equation}
 \vspace{-0.5em}
 \begin{equation}
     \sum\bold{C} = f(\bold{X}) 
 \end{equation}
where \(\odot\) denotes the Hadamard product. In the following subsections we provide detailed description of OMENN algorithm.

\subsection{Neural Network Representation}
Let \(f: \mathbb{R}^{d_0} \rightarrow \mathbb{R}^{d_n}\) represent a neural network that is composed of \(n\) layers (intermediate transformations \(l_1, ..., l_n\)):
\begin{equation}{\label{base_net}}
    f(\bold{x}) = l_n \circ l_{n-1} \circ ... \circ l_1(\bold{x})
\end{equation}
We assume that each layer \(l_i: \mathbb{R}^{d_{i-1}} \rightarrow \mathbb{R}^{d_i}\) can be formulated as an affine, or input-dependent affine transformation:

\begin{equation}{\label{layer_eq}}
    l_i(\bold{x}) = \bold{W}_i\bold{x} + \bold{b}_i
\end{equation}
where \(\bold{W}_i \in \mathbb{R}^{d_{i} \times d_{i-1}}\), \(\bold{b}_i \in \mathbb{R}^{d_i}\). By \textit{input-dependent}, we mean that both \(\bold{W}_i\) and \(\bold{b}_i\) are functions of \(\bold{x}\).

This assumption may seem restrictive at first, as it is typically associated with fully-connected layers, however its applications are much broader. In practice it can also represent normalization layers (such as batch norm or layer norm), convolutional layers, attention mechanisms, activation functions, and more (see Subsection~\ref{layers}) .

In our work we focus on the Vision Transformer architecture, which includes convolutional layer, attention mechanism~\cite{zhang2019self}, normalization layers~\cite{ioffe2015batch,lei2016layer}, fully-connected layers, and residual connections~\cite{he2016deep}. We also consider a Convolutional Neural Network architecture, which consists of similar layers but excludes the attention mechanism. Therefore, in the following sections we discuss all components of those two architectures.

\subsection{Layers as Affine Transformations}\label{layers}
In this subsection, we present how commonly used layers -- specifically fully-connected, scaled dot-product attention, and activation functions -- can be represented as affine transformations, such as in Eq.~\ref{layer_eq}. Note that similar formulations are defined for convolution, normalization, multi-head self-attention and residual connections in the Supplementary Materials.

Let us assume that a sequential input to a neural network layer is represented as a matrix \(\bold{X} \in \mathbb{R}^{t_{in} \times d_{in}}\) (for image data \(t_{in}\) represents spatial dimensions \(hw\), and \(d_{in}\) denotes the channel dimension \(c\)), and the corresponding output as a matrix \(\bold{Y} \in \mathbb{R}^{t_{out} \times d_{out}}\). To ensure consistent notation across different layers, we adopt a column-major vectorization of both the input and output of each layer.

\begin{equation}\label{vectorization}
\begin{split}
& \bold{x} = \text{vec}(\bold{X}) \rightarrow \bold{x} \in \mathbb{R}^{t_{in}d_{in}}\\
& \bold{y} = \text{vec}(\bold{Y})  \rightarrow \bold{y} \in \mathbb{R}^{t_{out}d_{out}}\\
\end{split}
\end{equation}
In the following paragraphs, we present the equation for each layer \(l\) in the form of \(\bold{Y} = l(\bold{X})\), and then convert it into a vectorized affine form \(\bold{y} = \bold{W}\bold{x} + \bold{b}\). We use \(\otimes\) to denote the Kronecker product, relying on its four key properties:
\begin{equation}\label{kron_add}
\bold{A} \otimes (\bold{B} + \bold{C}) = (\bold{A}\otimes \bold{B}) + (\bold{A} \otimes \bold{C})
\end{equation}
\begin{equation}\label{kron_mul}
(\bold{A} \otimes \bold{B})(\bold{C} \otimes \bold{D}) = (\bold{AC})\otimes(\bold{BD})
\end{equation}
\begin{equation}\label{vec_add}
\text{vec}(\bold{A} + \bold{B}) = \text{vec}(\bold{A}) + \text{vec}(\bold{B})
\end{equation}
\begin{equation}\label{vec_mul}
\text{vec}(\bold{ABC}) = (\bold{C}^{\top} \otimes \bold{A})\text{vec}(\bold{B})
\end{equation}

\paragraph{Fully-Connected.}\label{fc}
Fully-connected layers already apply an affine transformation to the input matrix \(\bold{X}^{\top}\):
\begin{equation}\label{unvec_fc}
    \bold{Y} = \bold{X}\bold{W}_l + \bold{B}_l
\end{equation}
\noindent where \(\bold{W}_l \in \mathbb{R}^{d_{in} \times d_{out}}\), \(\bold{B}_l \in \mathbb{R}^{t_{in} \times d_{out}}\).
To derive the equivalent result for the vectorized input \(\bold{x}\), the above equation takes the following form:
\begin{equation}\label{fc_vec}
\begin{split}
    \bold{y} &= \text{vec}(\bold{Y}) \overset{\ref{vec_add}}{=} \\
    &= \text{vec}({\bold{XW}_l}) + \text{vec}(\bold{B}_l) = \\
    &= \text{vec}({\bold{IXW}_l}) + \text{vec}(\bold{B}_l) \overset{\ref{vec_mul}}{=} \\
    &= \underbrace{(\bold{W}_l^{\top} \otimes \bold{I})}_{\bold{W}}\underbrace{\text{vec}(\bold{X})}_{\bold{x}} + \underbrace{\text{vec}(\bold{B}_l)}_{\bold{b}} = \\
    &= \bold{Wx} + \bold{b}
\end{split}
\end{equation}

\paragraph{Scaled Dot-Product Attention.} The attention operation can be seen as an input-dependent affine transformation:
\begin{equation}\label{attention}
    \bold{Y} = \underbrace{\text{softmax}\left(\frac{\bold{QK}^{\top}}{\sqrt{d_k}}\right)}_{\bold{A}}\bold{V} = \bold{AV}\
\end{equation}

\noindent where \(\bold{Q}\), \(\bold{K}\), and \(\bold{V}\), are results of fully-connected layers applied to \(\bold{X}\), and \(d_k\) is the scaling factor.
This equation directly converts to a vectorized form as:
\begin{equation}\label{vec_attn_1}
\begin{split}
    \bold{y} &= \text{vec}(\bold{Y}) = \\
    &= \text{vec}(\bold{AV}) = \text{vec}(\bold{AVI}) \overset{\ref{vec_mul}}{=} \\
    &= (\bold{I}^{\top}\otimes\bold{A})\text{vec}(\bold{V})
\end{split}
\end{equation}
As \(\bold{V}\) is obtained via fully-connected layer described in the previous paragraph, we apply Eq.~\ref{fc_vec}:
\begin{equation}\label{vec_attn}
\begin{split}
    \bold{y} &= (\bold{I}^{\top}\otimes\bold{A})\underbrace{((\bold{W}_v^{\top} \otimes \bold{I})\bold{x} + \text{vec}(\bold{B}_v))}_{\text{vec}(\bold{V})} = \\
    &= (\bold{I}^{\top}\otimes\bold{A})(\bold{W}_v^{\top} \otimes \bold{I})\bold{x} + (\bold{I}^{\top}\otimes\bold{A})\text{vec}(\bold{B}_v) \overset{\ref{kron_mul}}{=} \\
    &= \underbrace{(\bold{W}_v^{\top} \otimes \bold{A})}_{\bold{W}}\bold{x} + \underbrace{\text{vec}(\bold{AB}_v)}_{\bold{b}} = \bold{Wx} + \bold{b}
\end{split}
\end{equation}

\noindent where \(\bold{V} = \bold{X}\bold{W}_v + \bold{B}_v\).

\paragraph{Dynamic Linear Activation Functions.}
Modern neural networks utilize activation functions, such as GELU (Gaussian Error Linear Unit)~\cite{hendrycks2016gaussian} and SWISH (Self-Gated Activation Funstion)~\cite{Ramachandran2017SwishAS}, due to their superior performance in capturing complex patterns and enhancing model expressiveness. Both functions share a common structure:

\begin{equation}{\label{act}}
    \phi(x) = x \cdot \xi(x)
\end{equation}
For SWISH \(\xi(x) = \sigma(x)\), where \(\sigma\) is the sigmoid activation function, while for GELU \(\xi(x) = \Phi(x)\), where \(\Phi\) is the Cumulative Distribution Function (CDF) of a standard Gaussian distribution.
This input-dependent linear form allows us to directly formulate the vectorized form as:
\begin{equation}\label{act_vec}
\begin{split}
\bold{y} &= \bold{x} \odot \xi(\bold{x}) + \bold{0} = \\
&= \underbrace{\text{diag}(\xi(\bold{x}))}_{\bold{W}}\bold{x} + \underbrace{\bold{0}}_{\bold{b}} = \\
&= \bold{W}\bold{x} + \bold{b}
\end{split}
\end{equation}

\subsection{Neural Network as Affine Transformation}
In the aforementioned subsection, we showed that multiple layers employed in modern neural network architectures can be reformulated as affine (or input-dependent affine) transformations. Since a sequence of affine transformations can be equivalently represented as a single affine transformation, the output of \(f\) from Eq.~\ref{base_net} can be expressed as:

\begin{equation}{\label{omenn_net}}
\begin{split}
    f(\bold{x}) &= l_n \circ l_{n-1} \circ ... \circ l_1(\bold{x}) = \\
    &= \bold{\Omega_w}\bold{x} + \sum_{i=1}^n\bold{\Omega_{b_i}}\bold{b_i}
\end{split}
\end{equation}
where:
\begin{equation}\label{omega_w}
    \bold{\Omega_w} = \prod_{i=n}^1\bold{W}_i
\end{equation}

\begin{equation}\label{omega_b}
    \bold{\Omega_{b_i}} = \begin{cases}\prod_{j=n}^{i+1}\bold{W}_j, & i < n \\\\ \bold{I}_{t_nd_n}, & i = n\end{cases}
\end{equation}
Representation from Eq.~\ref{omenn_net} enables the entire neural network to be described by a single input-dependent affine transformation, explicitly revealing \textit{the direct contribution} \(\bold{\Omega_w}\) of each element of the input \(\bold{x}\), as well as \textit{the direct contribution} \(\bold{\Omega_{b_i}}\) of each element of the bias \(\bold{b_i}\), to the output score. 
By multiplying the inputs and biases by their respective direct contributions and summing the results, we obtain the exact output that the neural network \(f\) produces for the input \(\bold{x}\).

\subsection{Explaining Layers}\label{fusion}
In the previous subsection, we demonstrated how the neural network can be reformulated as a single input-dependent affine transformation, highlighting the direct contribution of the input \(\bold{x}\) and each bias term \(\bold{b_i}\) to the network's output.
However, this representation treats each bias \(\bold{b_i}\) as an independent input -- similar to \(\bold{x}\) -- rather than as part of the additive contribution derived from \(\bold{x}\). 

To solve the above problem we convert each layer, from input-dependent affine, to an equivalent, input-dependent linear form.
For this purpose, we first add an additional channel of ones \(\bold{1} \in \{1\}^{t_{in} \times 1}\) to the input \(\bold{X}\) resulting in the augmented \(\bold{\tilde{X}} \in \mathbb{R}^{t_{in} \times (d_{in}+1)}\):
\begin{equation}\label{input_aug}
    \bold{\tilde{X}} = \begin{bmatrix} \bold{X} & \bold{1} \end{bmatrix}
\end{equation}
Moreover, we combine the weights and the biases of each layer, what in the case of fully-connected layer from Subsection~\ref{fc} results in:
\begin{equation}\label{merge_fc}
    \bold{\tilde{W}}_l = \begin{bmatrix} \bold{W}_l & \bold{0} \\ \bold{b}_l & 1 \end{bmatrix}
\end{equation}
where \(\bold{b}_l \in \mathbb{R}^{1 \times d_{in}}\) is a row of \(\bold{B}_l = \bold{1}_{t_{in \times 1}}\bold{b}_l\) from Eq.~\ref{unvec_fc} and \(\bold{0} \in \{0\}^{(t_{out} - 1 )\times 1}\).
The output is then defined as:
\begin{equation}
    \bold{\tilde{Y}} = \bold{\tilde{X}\tilde{W}}_l = \begin{bmatrix} \bold{X}\bold{W}_l + \bold{B}_l & \bold{1} \end{bmatrix} = \begin{bmatrix} \bold{Y} & \bold{1} \end{bmatrix}
\end{equation}
Therefore, according to the Eq.~\ref{fc_vec}, the vectorized version of a fully-connected layer is given as:
\begin{equation}\label{fused_fc_vec}
    \bold{\tilde{y}} = \text{vec}(\bold{\tilde{X}\tilde{W}}_l) = \underbrace{(\bold{\tilde{W}}_l^{\top} \otimes \bold{I})}_{\bold{\tilde{W}}}\underbrace{\text{vec}(\bold{\tilde{X}})}_{\bold{\tilde{x}}} = \bold{\tilde{W}\tilde{x}}
\end{equation}
This results in the same outcome as in Eq.~\ref{fc_vec}, but does not contain any additive bias terms.

Note that similar operations are defined for other layers, such as convolution and normalization, in Supplementary Materials.

\subsection{Explaining Neural Network}\label{final_explanation}After combining weights and biases for each layer, the affine transform of the neural network $f(\mathbf{x})$ from Eq.~\ref{omenn_net} takes the following input-dependent linear form:
\begin{equation}\label{omenn_net_fused}
    f(\bold{x}) = \bold{\tilde{\Omega}_w\tilde{x}}
\end{equation}
\noindent where:
\begin{equation}
    \bold{\tilde{\Omega}_w} = \prod_{i=n}^1 \bold{\tilde{W}}_i
\end{equation}
This leads to a single contribution matrix \(\bold{\tilde{\Omega}_w}\) applied to the augmented input \(\bold{\tilde{x}}\), effectively resolving the issue of multiple contribution matrices, each with a different shape.

Since we are focused on the contribution to a single logit value, we assume \(f: \mathbb{R}^{t_{in} \times d_{in}} \rightarrow \mathbb{R}\). Therefore, the above components in a non-vectorized form are defined as follows: \(\bold{\tilde{X}} = \text{vec}^{-1}(\bold{\tilde{x})}\) and \(\bold{\tilde{C}_{wb}} = \text{vec}^{-1}(\bold{\tilde{\Omega}_w})\), where \(\bold{\tilde{X}},\bold{\tilde{C}_{wb}} \in \mathbb{R}^{t_{in} \times d_{in+1}}\).
Notably, the last input channel \(\bold{\tilde{X}}[:, d_{in}+1]\), composed of ones, corresponds to \(\bold{C_b}=\bold{\tilde{C}_{wb}}[:, d_{in}+1]\), representing the values added to the original input \(\bold{X}\). On ther other hand, \(\bold{C_w}=\bold{\tilde{C}_{wb}}[:, 1:d_{in}]\) represents the values that are multiplied with \(\bold{X}\).
This enables us to define the final explanation matrix \(\bold{C}\) where each element quantifies the contribution of the corresponding pixel to the overall score:
\begin{equation}\label{final_c}
    \bold{C} = \sum_{d}(\bold{\tilde{C}_{wb}} \odot \bold{\tilde{X}})[:, d] = \sum_{d}( \bold{C_w} \odot \bold{X})[:, d] + \bold{C_b}
\end{equation}
Additionally, it satisfies the completeness property \cite{tan2024post}:
\begin{equation}
\begin{split}
    \sum\bold{C} &= \sum (\bold{\tilde{C}_{wb}} \odot \bold{\tilde{X}}) = \bold{\tilde{\Omega}_w}\bold{\tilde{x}} = f(\bold{x})
\end{split}
\end{equation}

\noindent This compact representation not only preserves the interpretability of the weights and biases but also provides a clear decomposition of their roles in the network’s computations. The weights \(\bold{C_w}\) capture the multiplicative influence of the input, while the biases \(\bold{C_b}\) encapsulate the additive contributions.

\subsection{Relation to Gradient}\label{omenn_vs_grad}

To demonstrate the precision of OMENN as an attribution-based method, we compare its behavior to gradient approach. For linear functions (not input-dependent), both methods give identical results. However, this equivalence does not hold when evaluating non-linear functions, such as GELU.

\begin{figure}
    \centering
    \includegraphics[width=0.8\columnwidth]{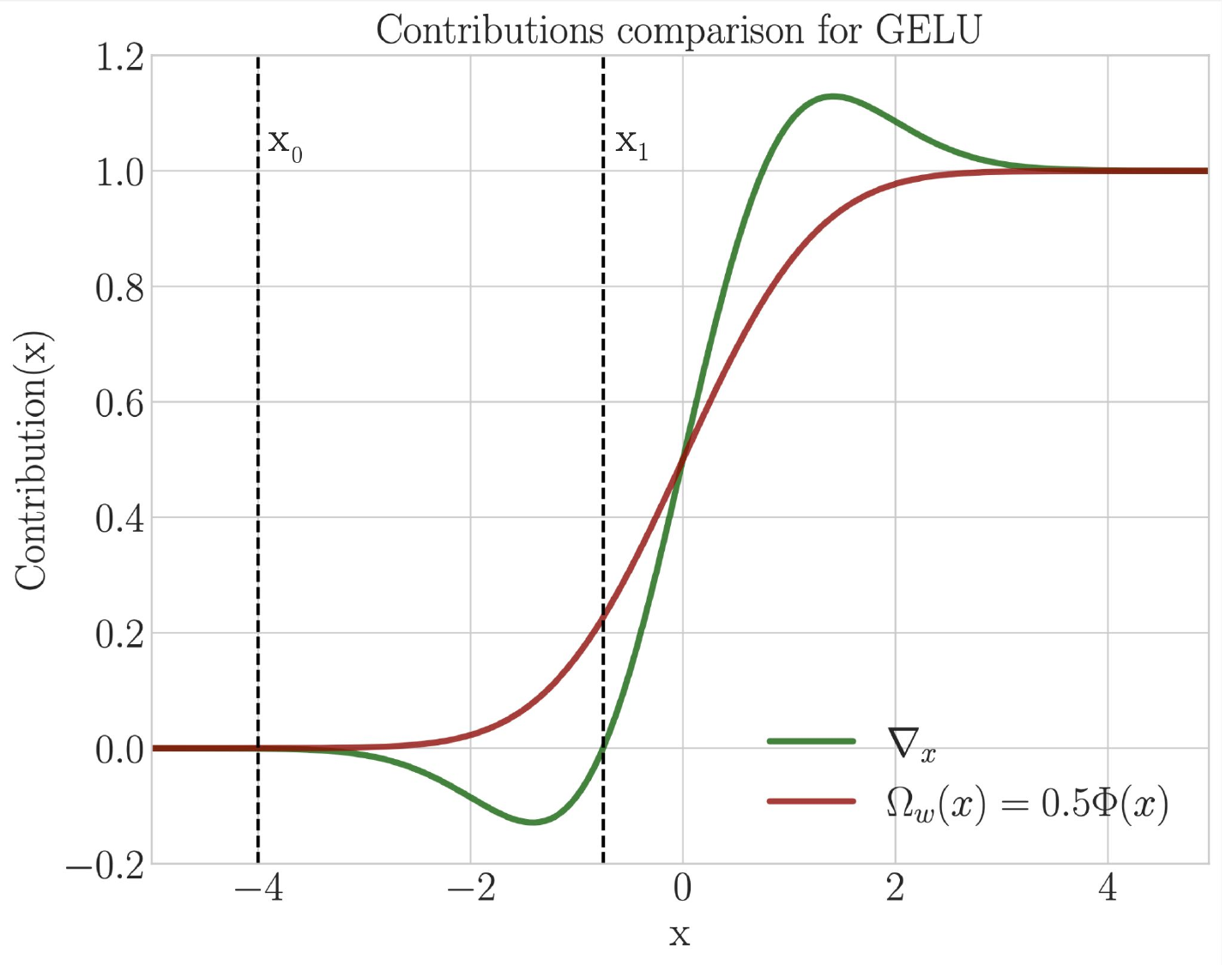}
    \begin{tabular}{l|ll}
        & $x_0= - 4$ & $x_1 = -0.75$ \\
        \hline
      $\nabla_x$ & 0 & 0 \\ 
      $\Omega_w(x)$ & 0 & 0.22 \\
      $\nabla_{x} \cdot x$ & 0 & 0 \\
      $\Omega_w(x) \cdot x $ & 0 & -0.17 \\
        ${GELU(x)} $ & 0 & -0.17 
    \end{tabular}
    \caption{An example highlighting the difference between our OMENN ($\Omega_w(x)$, red) and gradient ($\nabla_x$, green) methods for GELU. One can observe that gradient incorrectly returns the same contribution ($0$) for both considered points ($x_0=-4$ and $x_1=-0.75$). While our OMENN accurately provides higher contribution for $x_1$, which when multiplied by $x_1$ is equal to the output of GELU ($-0.17$), in contrast to output  obtained from the gradient ($0$).}
    \label{fig:gelu_grad_comp}
\end{figure}

The example of such situation is presented in Figure~\ref{fig:gelu_grad_comp} where we calculate the contribution using gradient and OMENN for a GELU activation function.
This example demonstrates that OMENN and Gradient methods can sometimes yield identical contributions—for instance, at $x_0=-4.0$. However, this is not always true, like for $x_1=-0.75$, where GELU returns a value of $-0.17$, while the partial derivative of GELU with respect to $x_1$ is $0$. This indicates that the Gradient method fails to accurately capture the underlying contribution of $x_1$.

In contrast, OMENN assigns a contribution of $0.22$ for $x_1$ and $0.0$ for $x_0$, effectively reflecting their true influence on the output. This results in a more reliable representation of each variable's contribution.

\section{Experimental Setup}
\paragraph{Datasets.} We evaluate our solution on synthetic dataset FunnyBirds~\cite{hesse2023funnybirds} used for benchmarking of explainable AI methods. We also use ImageNet1k~\cite{deng2009imagenet} for visualizations of the explanations produced by our method and to calculate the faithfulness metric~\cite{10.5555/3491440.3491857}.

\paragraph{FunnyBirds Benchmark Setup.}
The FunnyBirds dataset is made up of synthetically generated bird images, each created by combining five distinct, human-interpretable features: beak, wings, feet, eyes, and tail, referred to as "parts." It includes 50 bird categories, with each class representing a unique subset of 26 predefined parts. Additionally, the training set includes augmented images with missing parts.

The benchmark evaluates the methods on five different dimensions: \textbf{Accuracy (Acc.)} -- percentage of correct predictions on the test dataset.
\textbf{Background Invariance (B.I.)} -- Insensitivity of the model to background objects.
\textbf{Completness (Com.)} -- Explanation should highlight all relevant parts and removing parts identified as important should result in a different prediction.
\textbf{Correctness (Cor.)} -- Estimated importance of each part should be correlated to actual importance.
\textbf{Contrastivity (Con.)} -- Explanations for different classes should highlight class-specific part.

\begin{figure*}[t]
    \centering
    \includegraphics[width=0.95\textwidth]{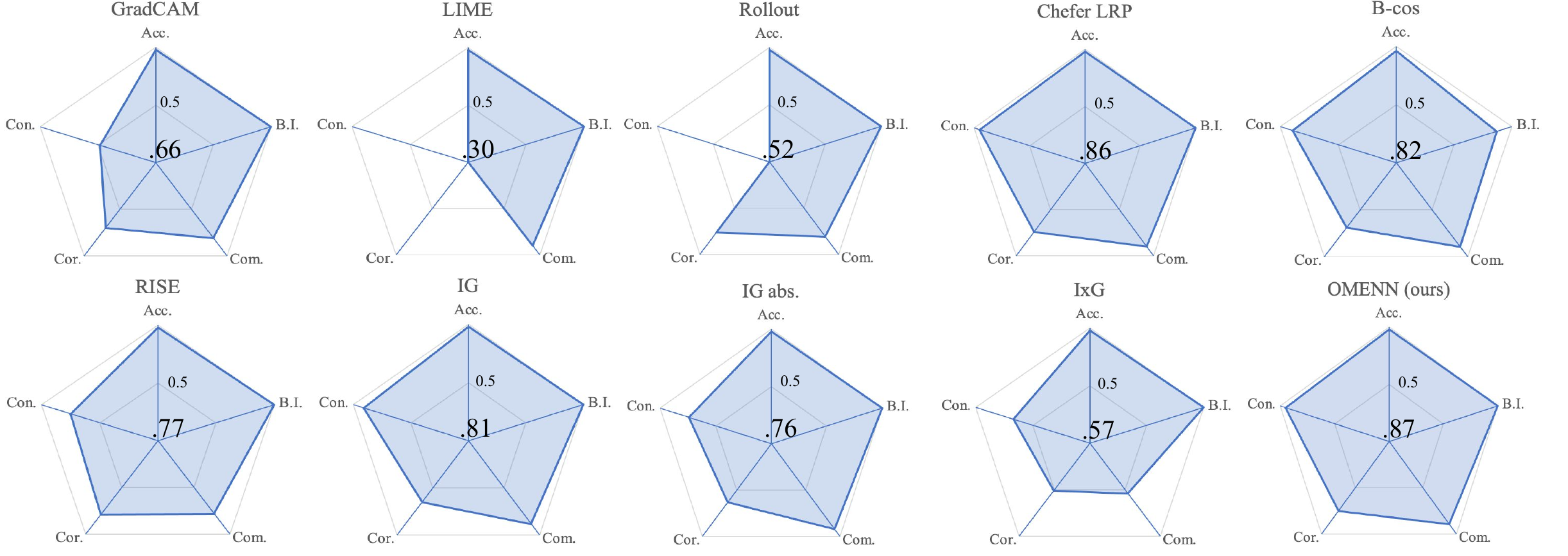}
    \caption{Results on the \textit{FunnyBirds} benchmark for the ViT-B/16 backbone show that OMENN achieves the highest score among various explainability methods. Notably, Chefer LRP ranks as the second-best approach, with only a marginal difference from OMENN, indicating that these methods are closely competitive.}
    \label{fig:fb_results}
\end{figure*}

\paragraph{Faithfulness.} We measure the Faithfulness Correlation metric~\cite{10.5555/3491440.3491857}, using Quantus~\cite{hedstrom2023quantus}, which measures the fidelity of model explanations by calculating the Pearson correlation between predicted logits and explanation attributions after randomly replacing a subset of features with baseline values. In our case the baseline value is 0.


\paragraph{Models.} To benchmark OMENN on the FunnyBirds framework, we use models provided by the framework's authors and apply OMENN to these models. Specifically, we employ the ViT-B/16 architecture~\cite{dosovitskiy2021imageworth16x16words} and the VGG-16 architecture~\cite{simonyan2015deepconvolutionalnetworkslargescale}, both pretrained on the FunnyBirds dataset. For evaluating the faithfulness metric, we use models pretrained on ImageNet, again ViT-B/16 and VGG-16.

\vspace{-1em}

\paragraph{Efficient Implementation of OMENN-based Explanations.}

We make the code publicly available\footnote{The code will be published after paper acceptance.}.

As described in Subsection~\ref{fusion}, we add an additional channel of ones to each input image \(\bold{X}\). This additional channel serves as a placeholder for the bias representation. 

To compute the weight and bias representation matrices \(\bold{C_w}\) and \(\bold{C_b}\), we temporarily modify the model's backward propagation. First, we convert all layers in the model to their combined parameter form (see Subsection \ref{fusion}). Next, we treat each tensor multiplied by the input as a constant by detaching it from gradient computations. Using the modified backward pass (with constant multipliers), we calculate the raw weights \(\bold{C_w}\) and biases \(\bold{C_b}\) representations as a single tensor (see Subsection \ref{final_explanation}). Finally, the contribution matrix \(\bold{C}\) is computed according to Eq. \ref{final_c}.

To focus exclusively on positive contributions, we disregard all elements with negative contributions.
Additionally, we perform post-processing steps, including quantile-based outlier removal and smoothing using a mean filter.

\section{Results}

\paragraph{\textit{FunnyBirds} Benchmark Comparison.} Figure~\ref{fig:fb_results} presents the results of OMENN on the \textit{FunnyBirds} framework, demonstrating that OMENN not only outperforms all other methods but also remains competitive with Chefer LRP. OMENN shows a marked improvement over other gradient-based approaches for assessing input image importance, highlighting its robustness. Furthermore, OMENN surpasses B-Cos, an inherently interpretable method, indicating that a precise explainability technique can effectively balance the trade-off between interpretability and model accuracy.

Supplementary Table~\ref{tab:fb_results} provides detailed numerical results across various backbones, showing that OMENN’s superiority is especially pronounced on the ViT backbone. This advantage is likely due to ViT's more complex architecture, particularly its non-linear elements, such as GELU, compared to VGG. Additionally, OMENN achieves the highest score in \textit{Contrastivity}, excelling in identifying image regions that drive distinctions between data classes. OMENN also scores well in \textit{Distractability}, indicating that its explanations avoid highlighting irrelevant parts of the input. However, OMENN performs less favorably on the \textit{Preservation Check} when compared to LRP and B-Cos, meaning that removing unimportant parts from the image does not always maintain the same prediction. Visual comparisons of results for VGG-16 across different methods are provided in the Supplementary Materials.

\begin{table}[]
    \centering
    \begin{tabular}{l|c}
    \toprule
         \textbf{Method} & \textbf{Faithfulness Score $\uparrow$}  \\
         \hline
         Random & $-0.004 \pm 0.004$ \\
         \hline
         Gradient & $0.032 \pm 0.008$\\
         GradCAM & $0.017 \pm 0.009$\\
         Integrated Gradient & $0.025 \pm 0.006$\\
         Chefer LRP & $0.022 \pm 0.008$\\
         OMENN (ours) &  $\mathbf{0.053 \pm 0.017}$\\
         \bottomrule
    \end{tabular}
    \caption{Evaluation of the faithfulness score~\cite{10.5555/3491440.3491857} using the Quantus benchmark~\cite{hedstrom2023quantus} for commonly used attribution-based explanation methods on ViT-B\/16. Our OMENN approach achieves the highest faithfulness score, surpassing all other methods and doubling the score of the second-best result. Notably, explanations derived solely from gradients also perform well, most likely because the metric uses a perturbation-based approach, which is preferable for gradient approaches. Finally, we include a random explanation as a baseline.}
    \label{tab:faithfulness}
    \vspace{-1.5em}
\end{table}

\paragraph{Comparison of Faithfulness.} 
Table~\ref{tab:faithfulness} provides a comparative analysis of the faithfulness of OMENN's explanations relative to other popular XAI methods. In evaluations conducted on ViT-B/16, OMENN consistently outperforms all competing approaches, underscoring its strong capability to generate explanations that closely align with the model's actual decision-making process. This enhanced faithfulness suggests that OMENN is a more reliable choice for explainable AI, achieving nearly double the score of the second-best methods, Chefer LRP and Integrated Gradients. Results for convolutional model are provided in the Supplementary Materials.

Interestingly, explanations derived solely from gradients are also shown to be effective, likely due to the perturbation-based approach used in calculating the metric. Since gradients identify the regions of an image where changes have the most significant impact on the neural network’s output, they can yield surprisingly faithful insights, even in the absence of more complex interpretive mechanisms.

\begin{figure}[t]
    \centering
    \includegraphics[width=0.48\textwidth]{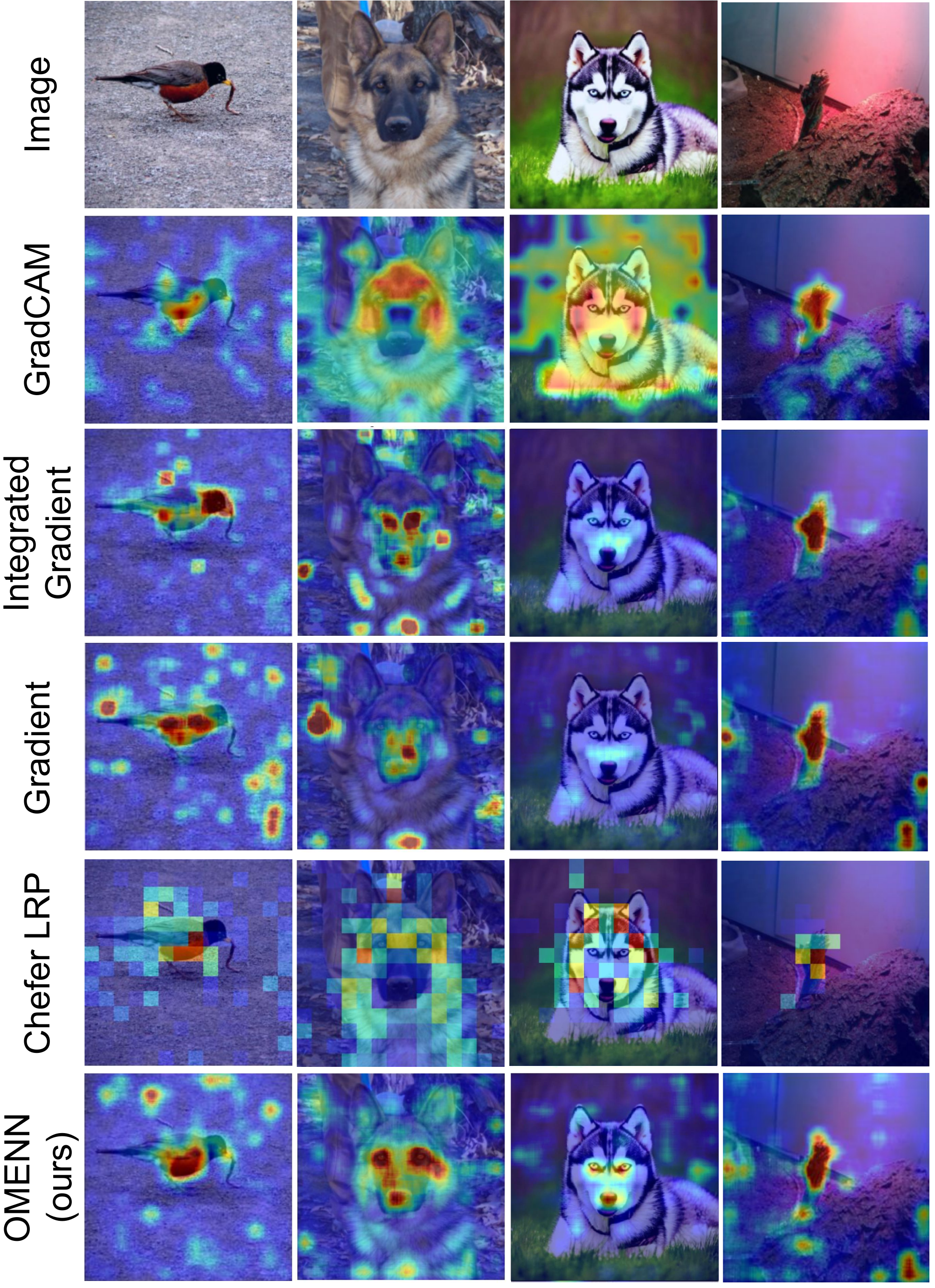}
    \caption{Examples of explanations generated by various XAI methods, including our OMENN approach, for ViT-B/16 on ImageNet. While OMENN’s contributions may appear noisy, they precisely reveal how each pixel influences the final logit value.}
    \label{fig:imnet_q}
\vspace{-1.25em}
\end{figure}

\paragraph{Qualitative Comparison.} Figure~\ref{fig:imnet_q} illustrates example explanations produced by various XAI techniques, demonstrating how OMENN and other methods allocate pixel importance in image classifications. OMENN consistently localizes the classified object effectively, concentrating importance on pixels tied to the object’s salient features. This focus contrasts with other methods, such as Gradient and GradCAM, which display less precise targeting. For instance, in the \textit{dog} example, OMENN focuses in on the dog's face, while Gradient and Integrated Gradient methods emphasize the background, and GradCAM and Chefer LRP distribute attention across the dog's general posture. Additionally, all of the XAI approaches tend to be noisy, only OMENN's explanations precisely reveal how each pixel influences the final logit value. More comparative examples are provided in the Supplementary Materials.

\section{Conclusions}

In this work, we propose OMENN, a method that provides locally exact explanations of predictions for modern neural networks, including CNNs and Vision Transformers (ViTs). We demonstrate both theoretically and experimentally that OMENN achieves state-of-the-art performance in explaining model predictions. In future work, we aim to explore how OMENN's computations can enhance processes such as knowledge distillation and continual learning by guiding regularization techniques.

\paragraph{Limitations.} The primary limitation of OMENN is that it cannot be applied to neural network architectures in which some operations cannot be formulated as an input-dependent affine transformation. Additionally, we do not address how plausible OMENN's explanations are for end-users, as this depends heavily on the method of visualization, which is beyond the scope of this work. Finally, we recognize that explanations may lead to overconfidence in model predictions~\cite{kim2022hive}, though mitigating this phenomenon is also outside the focus of this research.

\paragraph{Impact.} Our work primarily impacts the field of eXplainable Artificial Intelligence by demonstrating that neural network computations can be expressed in a locally linear form, both theoretically and experimentally. This advancement opens up new avenues of research into how such formulations could be applied, e.g. in knowledge distillation. Furthermore, this approach provides insights into the inner mechanisms of deep neural networks.

\section*{Acknowledgements}
The research was carried out for the project "Interpretable and Interactive Multimodal Retrieval in the Drug Discovery" (grant no. FENG.02.02-IP.05-0040/23) within the First Team FENG programme of the Foundation for Polish Science co-financed by the European Union under the European Funds for Smart Economy 2021-2027 (FENG). 

Some experiments were performed on servers purchased with funds from the Priority Research Area (Artificial Intelligence Computing Center Core Facility) under the Strategic Programme Excellence Initiative at Jagiellonian University.

We gratefully acknowledge Polish high-performance computing infrastructure PLGrid (HPC Center: ACK Cyfronet AGH) for providing computer facilities and support within computational grant no. PLG/2023/016555.


{
    \small
    \bibliographystyle{ieeenat_fullname}
    \bibliography{main}
}
\clearpage
\setcounter{page}{1}
\maketitlesupplementary

%

\section*{Mathematical Derivations}

\subsection*{Layers as affine transformations}
The aim of this subsection is to extend the work in subsection~\ref{layers} by reformulating the equations of commonly used layers into an affine transformation as described in Eq.~\ref{layer_eq} applied to the vectorized input defined in Eq.~\ref{vectorization}.

\paragraph{Normalization.} A typical normalization layer contains four parameters: \(\boldsymbol{\mu}\), \(\boldsymbol{\sigma}\), \(\boldsymbol{\gamma}\), \(\boldsymbol{\beta}\) \(\in \mathbb{R}^{d_{in}}\) and performs the following operation on the input $\bold{X}$:

\begin{equation}\label{norm}
    \bold{Y}[i,j] = \boldsymbol{\gamma}[j]\left(\dfrac{\bold{X}[i,j] - \boldsymbol{\mu}[j]}{\boldsymbol{\sigma}[j]}\right) + \boldsymbol{\beta}[j]
\end{equation}
We merge the above parameters into a scale \(\bold{s}_l \in \mathbb{R}^{d_{in}}\) and a bias \(\bold{b}_l \in \mathbb{R}^{d_{in}}\), which simplifies the normalization operation into an equivalent affine transformation:

\begin{equation}
    \bold{Y}[i,j] = \bold{X}[i,j]\bold{s}_l[j] + \bold{b}_l[j]
\end{equation}
where
\begin{equation}
    \bold{s}_l[j] = \left(\dfrac{\boldsymbol{\gamma}[j]}{\boldsymbol{\sigma}[j]}\right), \quad \bold{b}_l[j] = \left(\boldsymbol{\beta}[j] - \dfrac{\boldsymbol{\gamma}[j]\boldsymbol{\mu}[j]}{\boldsymbol{\sigma}[j]}\right)
\end{equation}
This operation can then be reformulated as a fully-connected layer, from Eq.~\ref{unvec_fc}:
\begin{equation}\label{norm_as_fc}
    \bold{Y} = \bold{X}\underbrace{\text{diag}(\bold{s}_l)}_{\bold{W}_l} + \bold{B}_l = \bold{X}\bold{W}_l + \bold{B}_l
\end{equation}
where \(\bold{B}_l = \bold{1}_{t_{in \times 1}}\bold{b}_l\). From this perspective, vectorization is obtained in the same way as for fully-connected layers from Eq.~\ref{fc_vec}.

\paragraph{Convolutional 2D.} 
For a convolutional layer, we reshape the input \(\bold{X} \in \mathbb{R}^{t_{in} \times d_{in}}\) into a tensor form \(\boldsymbol{\mathcal{X}} = \text{vec}^{-1}(\bold{X}, (h_{in}, w_{in}, d_{in}))\), where \(h_{in} \times w_{in}\) corresponds to the spatial dimensions, and \(d_{in}\) represents the number of input channels. The convolutional kernel \(\boldsymbol{\mathcal{K}} \in \mathbb{R}^{h_k \times w_k \times d_{in} \times d_{out}}\) is then applied to \(\boldsymbol{\mathcal{X}}\) via the cross-correlation operation \(\star\), followed by the addition of the bias \(\boldsymbol{\mathcal{B}} \in \mathbb{R}^{h_{out} \times w_{out} \times d_{out}}\). This produces the output \(\boldsymbol{\mathcal{Y}} \in \mathbb{R}^{h_{out} \times w_{out} \times d_{out}}\), where \(h_{out} \times w_{out}\) are the spatial dimensions of the output, and \(d_{out}\) is the number of output channels:
\begin{equation}\label{conv}
    \boldsymbol{\mathcal{Y}}[:, :, i] = \sum_{j=1}^{d_{in}}\boldsymbol{\mathcal{X}}[:, :, j] \star \boldsymbol{\mathcal{K}}[:, :, j, i] + \boldsymbol{\mathcal{B}}[:, :, i]
\end{equation}
We reshape \(\boldsymbol{\mathcal{Y}}\) into a matrix as:
\begin{equation}
    \bold{Y} = \text{vec}(\boldsymbol{\mathcal{Y}}, (t_{out}, d_{out}))
\end{equation}
where \(t_{out} = h_{out}w_{out}\). 
To express the convolutional layer as an affine transformation applied to the vectorized input, the doubly block-Toeplitz (DBT) matrix representation $\bold{T}(\boldsymbol{\mathcal{K}})$ can be utilized:
\begin{equation}\label{conv_vec}
\begin{split}
    \bold{y} &= \text{vec}(\bold{Y}) = \\ 
    &= \underbrace{\bold{T}(\boldsymbol{\mathcal{K}})}_{\bold{W}}\underbrace{\text{vec}(\bold{X})}_{\bold{x}} + \underbrace{\text{vec}(\boldsymbol{\mathcal{B}})}_{\bold{b}} =\\
    &= \bold{Wx} + \bold{b}
\end{split}
\end{equation}
This form of representing the convolutional layer is not commonly used due to the large size of \(\bold{T}(\boldsymbol{\mathcal{K}})\), which can lead to high computational cost. However, it provides a convenient way to reformulate the convolutional layer equation as an affine transformation applied to the vectorized input.

\paragraph{Piece-wise linear activation functions.} For piece-wise linear activation functions (like ReLU~\cite{krizhevsky2012imagenet} or Leaky ReLU~\cite{maas2013rectifier}), we use the following definition:
\begin{equation}\label{relu_act}
    \phi(x) = x \cdot \xi(x)
\end{equation}
where $\xi(x)$ is given by:
\begin{equation}
    \xi(x) = \begin{cases}1, & x > 0 \\ \alpha, & x \leq 0\end{cases}
\end{equation}
with \(\alpha = 0\) for ReLU. This formulation enables vectorization as shown in Eq.~\ref{act_vec}. Furthermore, since \(\xi(x)\) is equal to the derivative \(\nabla_{x}\phi(x)\), OMENN produces the same results as the gradient for piecewise linear activation functions.

\paragraph{Residual connections.} We define a generalized residual connection as an expression of the following form:  
\begin{equation}\label{residual}
    \bold{Y} = \sum_{i=1}^nf_i(\bold{X})
\end{equation}
Assuming that all layers within each \(f_i\) are affine (or input-dependent affine) transformations as described in previous sections, the above equation can be expressed as a single, unified affine transformation:  
\begin{equation}
\begin{split}
    \bold{y} &= \text{vec}(\bold{Y}) \overset{\ref{vec_add}}{=} \\
    &= \sum_{i=1}^n\text{vec}(f_i(\bold{X})) = \\
    &= \sum_{i=1}^n(\bold{W}_i\bold{x} + \bold{b}_i) = \\
    &= \underbrace{\sum_{i=1}^n\bold{W}_i}_{\bold{W}}\bold{x} + \underbrace{\sum_{i=1}^n\bold{b}_i}_{\bold{b}} = \\
    &= \bold{Wx} + \bold{b}
\end{split}
\end{equation}
Here, \(\bold{W}_i\) and \(\bold{b}_i\) are the weights and biases of the vectorized \(f_i\).

\paragraph{Multi-head Self Attention.} 
Multi-head attention uses \(h\) attention heads (see Eq. \ref{attention}) in parallel, then concatenates their result and applies the affine output projection via fully-connected layer.

\begin{equation}\label{mh_attn}
\begin{split}
    \bold{Y} &= \underbrace{\begin{bmatrix}\bold{A}_1\bold{V}_1 & \dots & \bold{A}_h\bold{V}_h\end{bmatrix}}_{\bold{C}}\bold{W}_u + \bold{B}_u 
\end{split}
\end{equation}
where \(\bold{A}_i\) are \(i_{th}\) attention head weights, \(\bold{V}_i\) are \(i_{th}\) attention head values, and \(\bold{W}_u, \bold{B}_u\) are output projection parameters.
We reformulate the above concatenation into the following vectorized form:

\begin{equation}
\begin{split}
\bold{c} &= \text{vec}(\bold{C}) = \begin{bmatrix}\text{vec}(\bold{A}_1\bold{V}_1) \\ \vdots \\ \text{vec}(\bold{A}_h\bold{V}_h)\end{bmatrix} \overset{\ref{vec_attn}}{=} \\
&= \begin{bmatrix}(\bold{W}_{v_1}^{\top} \otimes \bold{A}_1)\bold{x} + \text{vec}(\bold{A}_1\bold{B}_{v_1}) \\ \vdots \\ (\bold{W}_{v_h}^{\top} \otimes \bold{A}_h)\bold{x} + \text{vec}(\bold{A}_h\bold{B}_{v_h})\end{bmatrix} = \\
&= \underbrace{\begin{bmatrix}(\bold{W}_{v_1}^{\top} \otimes \bold{A}_1) \\ \vdots \\ (\bold{W}_{v_h}^{\top} \otimes \bold{A}_h)\end{bmatrix}}_{\bold{W}_c}\bold{x} + \underbrace{\begin{bmatrix}\text{vec}(\bold{A}_1\bold{B}_{v_1}) \\ \vdots \\\text{vec}(\bold{A}_h\bold{B}_{v_h}) \end{bmatrix}}_{\bold{b}_c} = \\
&= \bold{W}_c\bold{x} + \bold{b}_c
\end{split}
\end{equation}
As the output mapping is expressed via fully-connected layer, we proceed with its vectorization according to Eq.~\ref{fc_vec}.

\subsection*{Combining parameters}
The aim of this subsection is to expand subsection~\ref{fusion} by formulating a set of methods that merge the weights and biases for each layer, utilizing an input augmented with a channel of ones (Eq.~\ref{input_aug}). These methods ensure that the augmented channel is effectively incorporated into the computations and passed to the next layer in the network. Additionally, we reformulate each layer to a vectorized linear (or input-dependent linear) form.

\paragraph{Normalization.} As we have shown in Eq.~\ref{norm_as_fc}, normalization layers can be expressed as fully-connected layers. Following that conclusion, we combine normalization layer parameters according to Eq.~\ref{merge_fc}.

\paragraph{Convolutional 2D.} 
Convolutional layer (see Eq.~\ref{conv}) is the one that besides the feature dimension ${d_{in}}$ also changes the spatial dimension $t_{in}$. To accurately represent the bias addition as matrix multiplication, we follow the approach that uniformly distributes the bias addition onto the corresponding image patch (receptive field).

Using $im2col$ algorithm~\cite{yanai2016efficient, heide2015fast}, each element of the output of the convolutional layer can be expressed as a matrix multiplication between a local patch of the input image and a flattened version of the convolutional kernel:
\begin{equation}
\begin{split}
    \bold{Y}[i, :] &= \bold{K}\underbrace{im2col(\bold{X})[:, i]}_{\bold{x_{P_i}}} + \bold{b} = \bold{K}\bold{x_{P_i}} + \bold{b_P}
\end{split}
\end{equation}
where $\bold{K} = \text{vec}(\boldsymbol{\mathcal{K}}, (h_kw_kd_{in}, d_{out}))^{\top}$, $\bold{b_P} = \text{vec}(\boldsymbol{\mathcal{B}}, (t_{out}, d_{out}))[i, :]$ (note that for each $i$, $\bold{b_p}$ remains unchanged). For the augmented input image $\bold{\tilde{X}}$, the additional channel of ones is incorporated, transforming the patch-based representation as follows:
\begin{equation}
    \bold{\tilde{x}_{P_i}} = im2col(\bold{\tilde{X}})[:, i] = \begin{bmatrix}\bold{x_{P_i}} \\ \bold{c_{P_i}} \end{bmatrix}
\end{equation}
where $\bold{c_{P_i}} \in \{0, 1\}^{h_kw_k}$ represents the corresponding patch of the added channel of ones. Note that $\bold{c_{P_i}}$ may contain zeros due to zero-padding applied to the input image.
To combine convolutional parameters, we design two separate kernels: static $\bold{\tilde{K}_w}$ and dynamic $\bold{\tilde{K}_{b_i}}$, such that $\bold{\tilde{K}_w},\bold{\tilde{K}_{b_i}} \in \mathbb{R}^{(d_{out}+1) \times h_kw_k(d_{in}+1)}$:
\begin{equation}
\begin{split}
    \bold{\tilde{K}_w} = \begin{bmatrix} \bold{K} & \bold{0} \\ \\ \bold{0} & \bold{0} \end{bmatrix}, \quad
    \bold{\tilde{K}_{b_i}} = \dfrac{1}{s_i}\begin{bmatrix} \bold{0} & \bold{\tilde{B}} \\ \\ \bold{0} & \bold{1} \end{bmatrix}
\end{split}
\end{equation}
where $\bold{\tilde{B}}=\underbrace{\begin{bmatrix} \bold{b_P} & \bold{b_P} & \dots & \bold{b_P} \end{bmatrix}}_{h_kw_k \quad \text{times}}$ and $s_i=\sum\bold{c_{P_i}}$.$\qquad$ 
We compute the final result by summing the outcomes obtained from applying both kernels to the augmented input $\bold{\tilde{X}}$:
\begin{equation}
\begin{split}
    \bold{\tilde{Y}}[i, :] &= \bold{\tilde{K}_w}\bold{\tilde{x}_{P_i}} +  \bold{\tilde{K}_{b_i}}\bold{\tilde{x}_{P_i}} = \\
    &= \begin{bmatrix}\bold{Kx_{P_i}} \\ 0 \end{bmatrix} + \begin{bmatrix}\bold{b_P} \\ 1 \end{bmatrix} = \begin{bmatrix}\bold{Y}[i,:] \\ 1\end{bmatrix}
\end{split}
\end{equation}
Which further implies that $\bold{\tilde{Y}} = \begin{bmatrix}\bold{Y} & \bold{1} \end{bmatrix}$.
Similarly to Eq.~\ref{conv_vec}, the above expression can be vectorized using doubly block-Toeplitz matrix (DBT) representation:
\begin{equation}
\begin{split}
    \bold{\tilde{y}} &= \text{vec}(\bold{\tilde{Y}}) = \\ &= \bold{T}(\boldsymbol{\tilde{\mathcal{K}}_w})\bold{\tilde{x}} + \bold{T}(\boldsymbol{\tilde{\mathcal{K}}_{b_i}})\bold{\tilde{x}} = \\
    &= \underbrace{(\bold{T}(\boldsymbol{\tilde{\mathcal{K}}_w}) + \bold{T}(\boldsymbol{\tilde{\mathcal{K}}_{b_i}}))}_{\bold{\tilde{W}}}\bold{\tilde{x}} = \bold{\tilde{W}}\bold{\tilde{x}}
\end{split}
\end{equation}
where $\boldsymbol{\tilde{\mathcal{K}}_w} = \text{vec}^{-1}(\boldsymbol{\tilde{K}_w}^{\top}, (h_k, w_k, d_{in}+1, d_{out}+1))$, $\boldsymbol{\tilde{\mathcal{K}}_{b_i}} = \text{vec}^{-1}(\boldsymbol{\tilde{K}_{b_i}}^{\top}, (h_k, w_k, d_{in}+1, d_{out}+1))$.

\paragraph{Activation functions.}
For all of the considered activation functions, the bias parameter $\bold{b} = \bold{0}$, meaning that we do not need to combine it with the dynamic weight matrix. However we still need to pass the augmented channel of ones through the network. This is done by extending the multiplier $\xi(\bold{X})$ with a column of ones as follows:
\begin{equation}
\begin{split}
    \bold{\tilde{Y}} &= \bold{\tilde{X}} \odot \begin{bmatrix}\xi(\bold{X}) & \bold{1}\end{bmatrix} = \\
    &= \begin{bmatrix}\bold{X}\odot\xi(\bold{X}) & \bold{1}\end{bmatrix} =  \\
    &= \begin{bmatrix}\bold{Y} & \bold{1}\end{bmatrix} 
\end{split}
\end{equation}
Similarly to Eq.~\ref{act_vec} the vectorization of $\bold{\tilde{Y}}$ is formulated as:
\begin{equation}
    \begin{split}
        \bold{\tilde{y}} &= \text{vec}(\bold{\tilde{Y}}) = \\
        &= \underbrace{\text{vec}(\bold{\tilde{X}})}_{\bold{\tilde{x}}} \odot \underbrace{\text{vec}(\begin{bmatrix}\xi(\bold{X}) & \bold{1}_{t_{in}}\end{bmatrix})}_{\bold{\tilde{w}}} = \\
        &= \underbrace{\text{diag}(\bold{\tilde{w}})}_{\bold{\tilde{W}}}\bold{\tilde{x}} = \bold{\tilde{W}}\bold{\tilde{x}}
    \end{split}
\end{equation}

\paragraph{Scaled Dot-Product Attention.} For the attention mechanism (see Eq.~\ref{attention}), only the values mapping \(\bold{V}\) involves bias addition (we treat \(\bold{Q}\) and \(\bold{K}\) as components of \(\bold{A}\)). Since \(\bold{V}\) is obtained via fully-connected layer, we merge its parameters (weight \(\bold{W}_v\) and bias \(\bold{B}_v\)) into a combined weight \(\bold{\tilde{W}}_v\), similarly to Eq.~\ref{merge_fc}. Additionally, to pass the augmented channel of ones we define \(\bold{\tilde{W}}_{aug}\).
\begin{equation}\label{attn_combine}
    \bold{\tilde{W}}_v = \begin{bmatrix} \bold{W}_v \\ \bold{b}_v \end{bmatrix}, \quad \bold{\tilde{W}}_{aug} =  \begin{bmatrix} \bold{0}_{d_{in}} \\ 1 \end{bmatrix}
\end{equation}
We compute the augmented output $\bold{\tilde{Y}}$ as the following concatenation:
\begin{equation}
\begin{split}\label{merge_attn}
    \bold{\tilde{Y}} &= \begin{bmatrix}\bold{A\tilde{X}\tilde{W}}_v & \underbrace{\bold{\tilde{X}}\bold{\tilde{W}}_{aug}}_{\bold{1}}\end{bmatrix} = \\
    &= \begin{bmatrix} \bold{A}(\underbrace{\bold{XW}_v + \bold{B}_v}_{\bold{V}}) & \bold{1} \end{bmatrix} = \\
    &= \begin{bmatrix} \bold{AV} & \bold{1} \end{bmatrix} = \\
    &= \begin{bmatrix} \bold{Y} & \bold{1} \end{bmatrix}
\end{split}
\end{equation}
$\bold{\tilde{Y}}$ can be vectorized into the following form:
\begin{equation}\label{fused_attn_vec}
    \begin{split}
        \bold{\tilde{y}} &= \text{vec}(\bold{\tilde{Y}}) = \\
        &= \begin{bmatrix}\text{vec}(\bold{A\tilde{X}\tilde{W}}_v) \\ \text{vec}(\bold{\tilde{X}}\bold{\tilde{W}}_{aug})\end{bmatrix} \overset{\ref{vec_mul}}{=} \\
        &= \begin{bmatrix}(\bold{\tilde{W}}_v^{\top}\otimes \bold{A})\text{vec}(\bold{\tilde{X}}) \\ (\bold{\tilde{W}}_{aug} \otimes \bold{I})\text{vec}(\bold{\tilde{X}})\end{bmatrix} = \\
        &= \underbrace{\begin{bmatrix}\bold{\tilde{W}}_v^{\top}\otimes \bold{A} \\ \bold{\tilde{W}}_{aug} \otimes \bold{I}\end{bmatrix}}_{\bold{\tilde{W}}}\bold{\tilde{x}} = \bold{\tilde{W}}\bold{\tilde{x}}
    \end{split}
\end{equation}

\paragraph{Multi-head self attention.} For multi-head attention mechanism we proceed in a similar way to scaled dot-product attention (see Eq.~\ref{attn_combine}). At first we combine the bias of each values mapping $\bold{b}_{v_i}$ with its corresponding weight $\bold{W}_{v_i}$. Then we define an additional weight matrix $\bold{\tilde{W}}_{aug}$ to pass the augmented channel of ones:
\begin{equation}
    \bold{\tilde{W}}_{v_i} = \begin{bmatrix} \bold{W}_{v_i} \\ \bold{b}_{v_i} \end{bmatrix}, \quad \bold{\tilde{W}}_{aug} =  \begin{bmatrix} \bold{0}_{d_{in}} \\ 1 \end{bmatrix}
\end{equation}
This converts the concatenation $\bold{C}$ from Eq.~\ref{mh_attn} to the following augmented form:
\begin{equation}
\begin{split}
\bold{\tilde{C}} &= \begin{bmatrix}\bold{A}_1\underbrace{\bold{\tilde{X}}\bold{\tilde{W}}_{v_1}}_{\bold{V}_1} & \dots & \bold{A}_h\underbrace{\bold{\tilde{X}}\bold{W}_{v_h}}_{\bold{V}_h} & \underbrace{\bold{\tilde{X}}\bold{\tilde{W}}_{aug}}_{\bold{1}}\end{bmatrix} = \\
&= \begin{bmatrix}\bold{A}_1\bold{V}_1 & \dots & \bold{A}_h\bold{V}_h & \bold{1}\end{bmatrix} = \\
&= \begin{bmatrix}\bold{C} & \bold{1}\end{bmatrix}
\end{split}
\end{equation}
We proceed with vectorization of the above result similarly to Eq.~\ref{fused_attn_vec}:

\begin{equation}
    \begin{split}
        \bold{\tilde{c}} &= \text{vec}(\bold{\tilde{C}}) = \\
        &= \begin{bmatrix}\text{vec}(\bold{A}_1\bold{\tilde{X}\tilde{W}}_{v_1}) \\ \vdots \\
        \text{vec}(\bold{A}_h\bold{\tilde{X}\tilde{W}}_{v_h}) \\ \text{vec}(\bold{\tilde{X}}\bold{\tilde{W}}_{aug})\end{bmatrix} \overset{\ref{fused_attn_vec}}{=} \\
        &= \underbrace{\begin{bmatrix}\bold{\tilde{W}}_{v_1}^{\top}\otimes \bold{A} \\ \vdots \\
        \bold{\tilde{W}}_{v_h}^{\top}\otimes \bold{A} \\ \bold{\tilde{W}}_{aug} \otimes \bold{I}\end{bmatrix}}_{\bold{\tilde{W}}}\bold{\tilde{x}} = \bold{\tilde{W}}\bold{\tilde{x}}
    \end{split}
\end{equation}
As the output mapping is expressed via fully-connected layer (see Eq.~\ref{mh_attn}), we proceed with its augmented vectorization according to Eq.~\ref{fused_fc_vec}.

\section*{Visualizations}
\begin{figure}[!htbp]
    \centering
    \includegraphics[width=0.9\columnwidth]{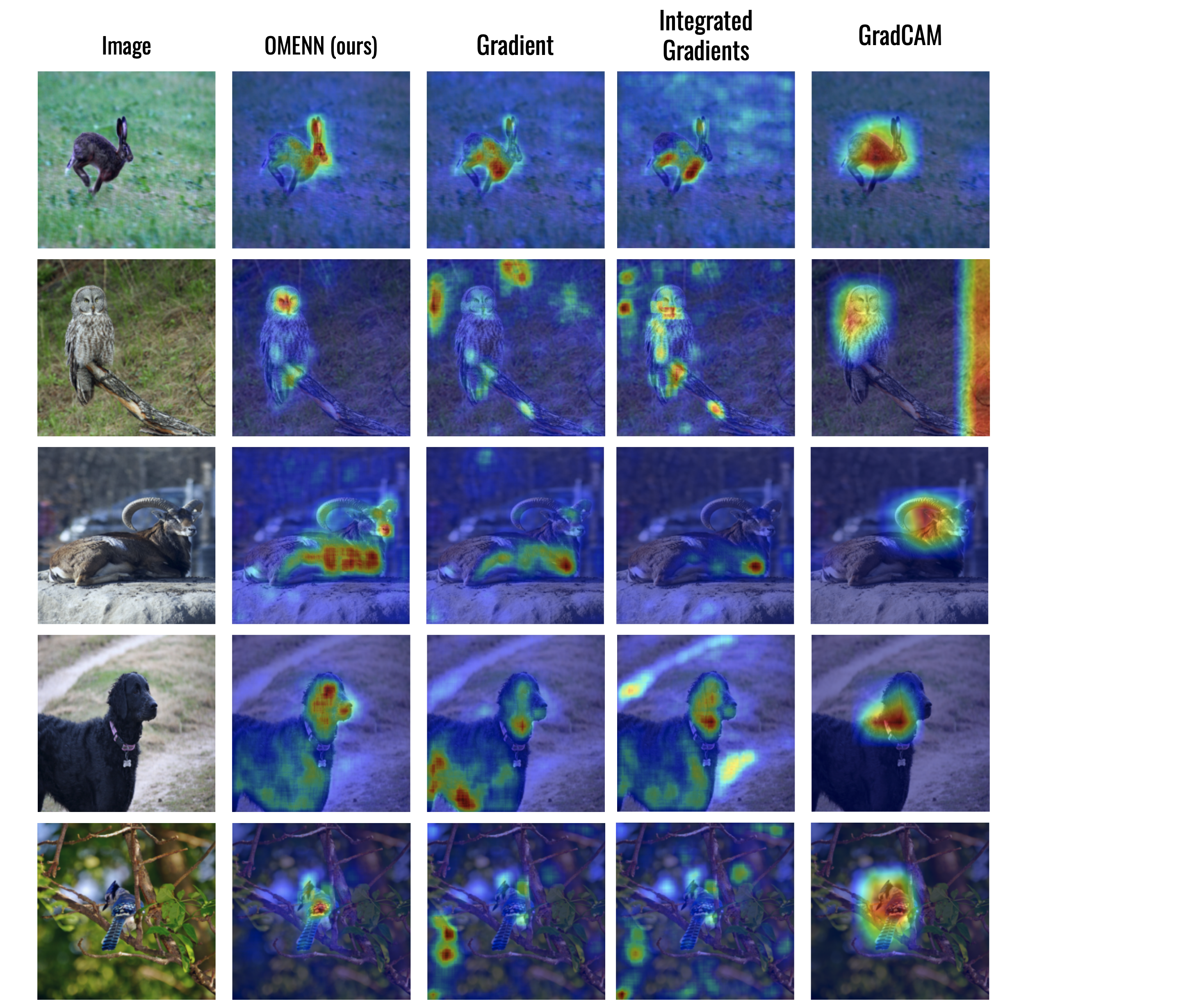}
    \caption{Examples of explanations generated by various XAI methods, including OMENN approach, for ConvNeXt Base.}
    \label{fig:convnext_mosaic}
\end{figure}

\begin{figure}[!htbp]
    \centering
    \includegraphics[width=0.9\columnwidth]{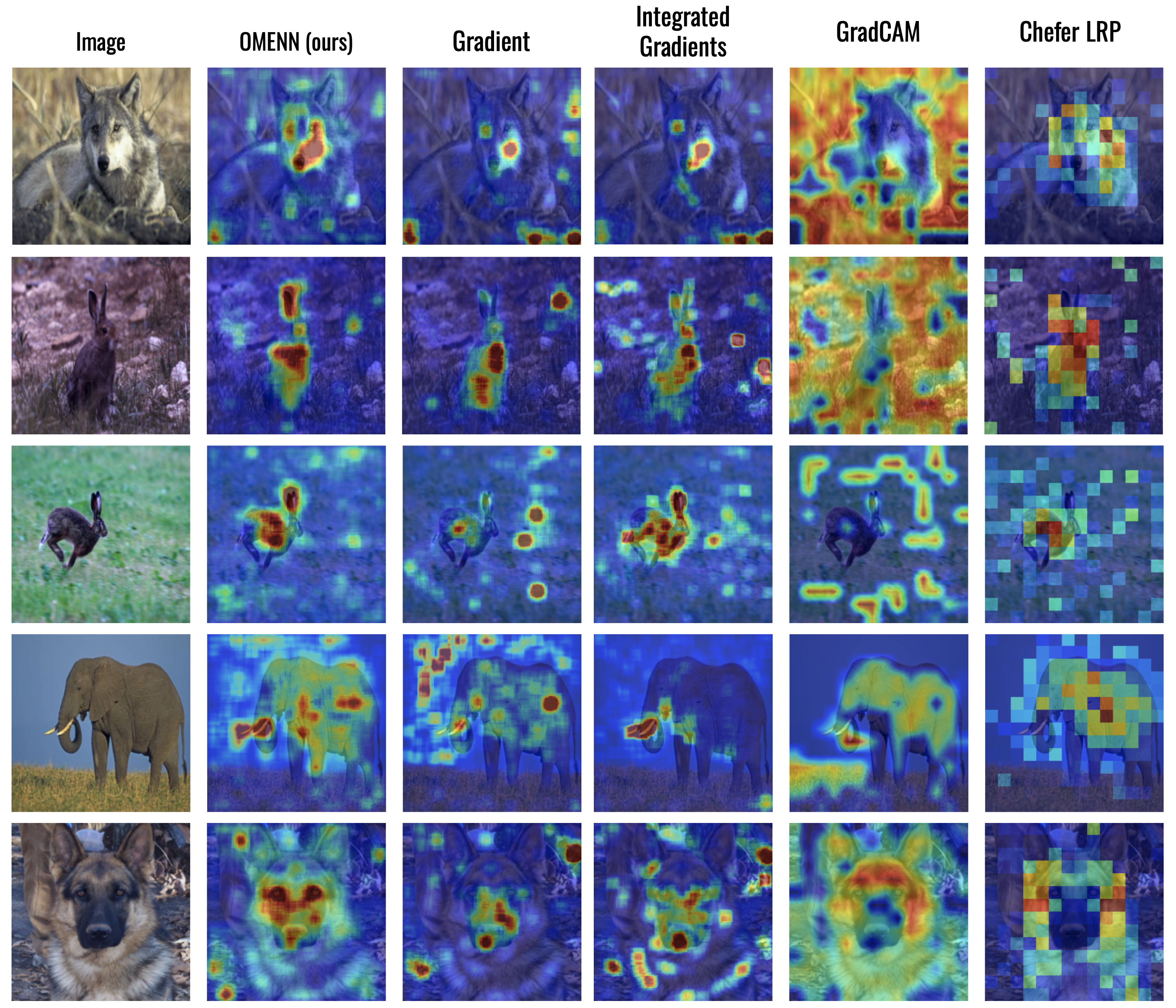}
    \caption{Examples of explanations generated by various XAI methods, including OMENN approach, for ViT-B/16.}
    \label{fig:vit_mosaic}
\end{figure}

\begin{figure}[!htbp]
    \centering
    \includegraphics[width=0.9\columnwidth]{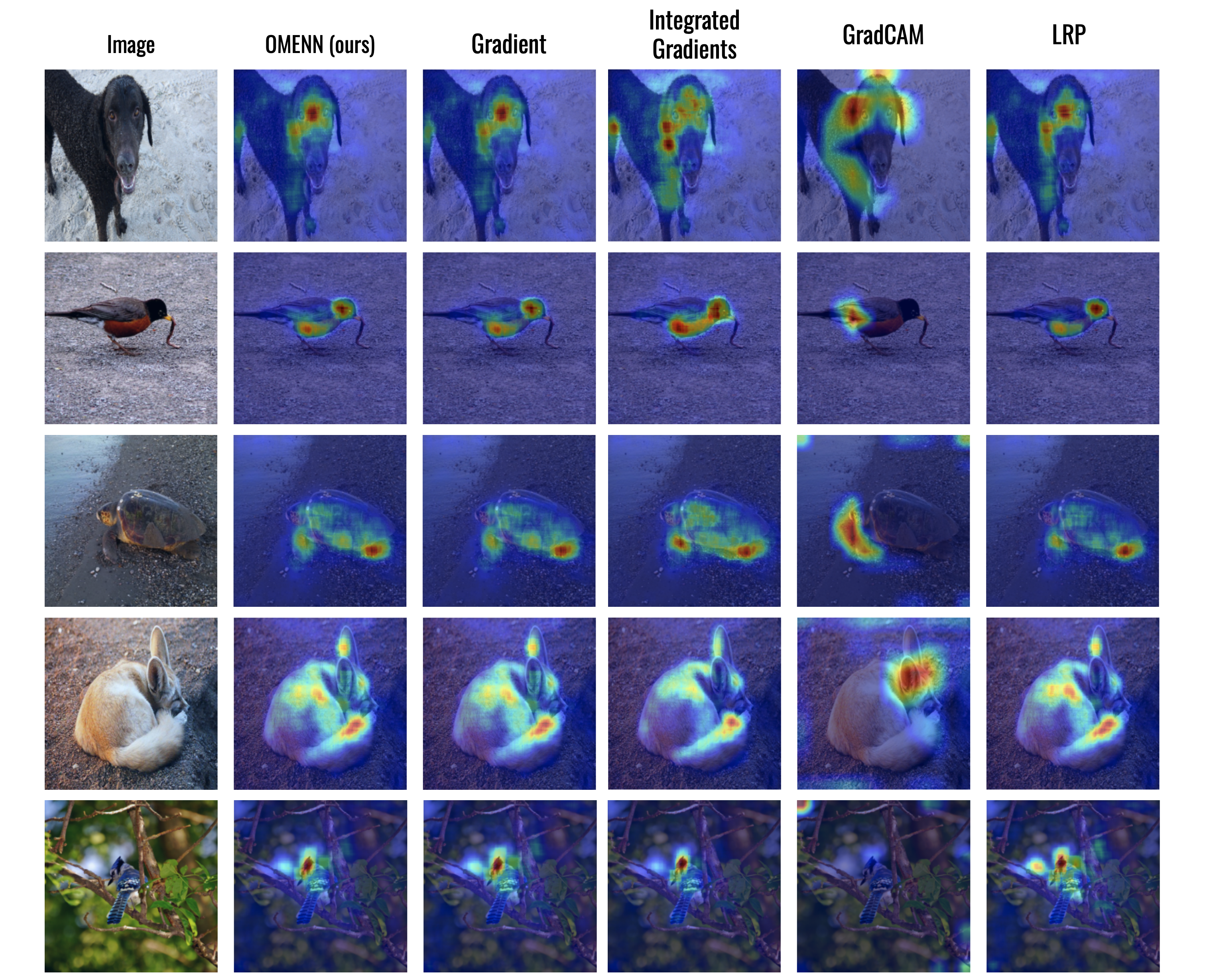}
    \caption{Examples of explanations generated by various XAI methods, including OMENN approach, for VGG-16. Note that according to Eq.~\ref{relu_act} OMENN explanation for VGG-16 is equivelent to image times gradient with additional bias contribution.}
    \label{fig:vgg_mosaic}
\end{figure}

In this section, we present contribution maps generated by OMENN, across several image classifiers (see Figures~\ref{fig:convnext_mosaic}-\ref{fig:vgg_mosaic}). These visualizations highlight the direct contributions of input pixels to the model's output. We compare the results with those produced by other widely used attribution methods. While OMENN’s contributions may appear noisy, they precisely reveal how each pixel influences the final logit value.

\section*{Experiments results}
\begin{figure*}[!htbp]
    \centering
    \includegraphics[width=0.9\textwidth]{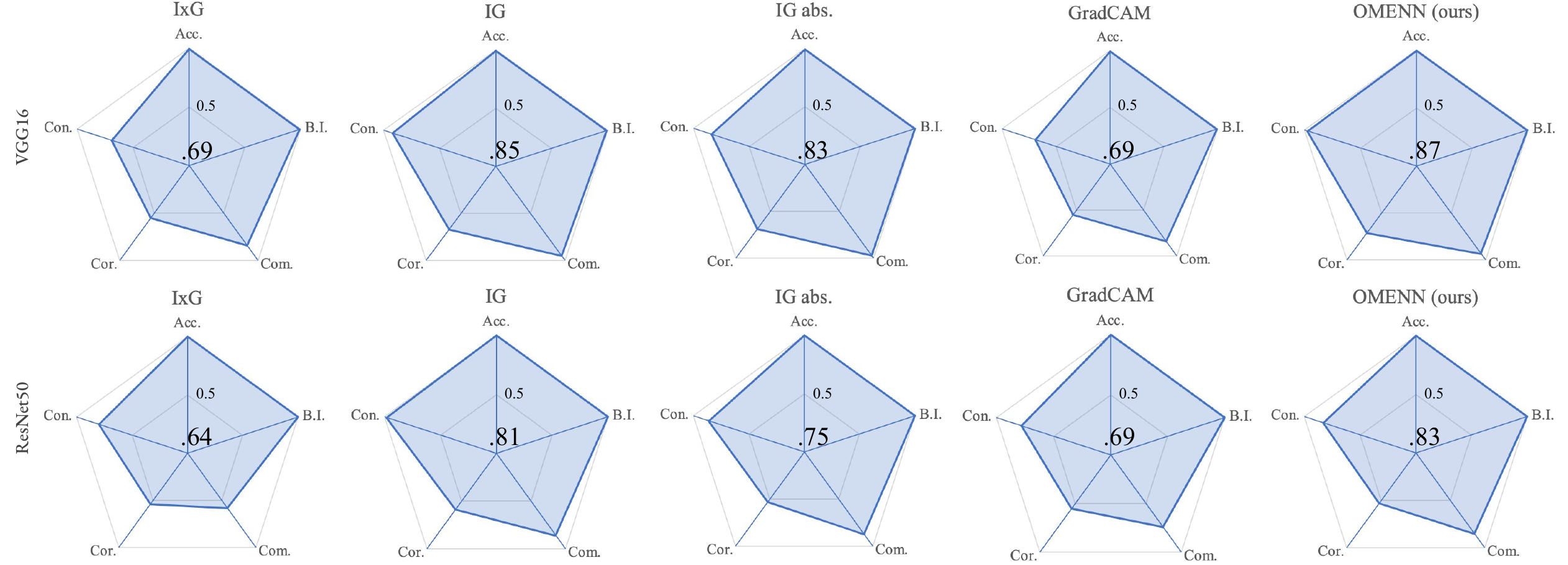}
    \caption{Results of OMENN on FunnyBirds framework using CNN backbones, VGG16 and ResNet50. OMENN surpasses all other XAI methods. Note that the choice of the backbone has a tremnedous influence on the XAI method performance, e.g. OMENN on ResNet50 achieves 5\% lower results than on VGG-16.}
    \label{fig:CNN_FB}
\end{figure*}

In this section, we present a comprehensive evaluation of OMENN and other widely-used attribution-based explanation methods across various benchmarks. Table~\ref{tab:faithfulness_vgg} reports the faithfulness score~\cite{10.5555/3491440.3491857} of different XAI methods using the Quantus benchmark~\cite{hedstrom2023quantus} on the VGG-16 backbone. Table~\ref{tab:fb_results} summarizes the performance of OMENN and other methods within the FunnyBirds framework using ViT-16/B, ResNet-50, and VGG-16 backbones. Additionally, Figure~\ref{fig:CNN_FB} highlights the performance of OMENN on the FunnyBirds framework with CNN backbones, demonstrating its superior results compared to competing methods.
\begin{table}[!htbp]
    \centering
    \begin{tabular}{l|c}
    \toprule
         \textbf{Method} & \textbf{Faithfulness Score $\uparrow$}  \\
         \hline
         Random & $-0.006 \pm 0.004$ \\
         \hline
         Gradient & $0.0214 \pm 0.007$\\
         GradCAM & $0.065 \pm 0.022$\\
         Integrated Gradient & $0.045 \pm 0.008$\\
         LRP & $0.057 \pm 0.002$\\
         OMENN (ours) &  $0.023 \pm 0.003$\\
         \bottomrule
    \end{tabular}
    \caption{Evaluation of the faithfulness score~\cite{10.5555/3491440.3491857} using the Quantus benchmark~\cite{hedstrom2023quantus} for commonly used attribution-based explanation methods on VGG16.}
    \label{tab:faithfulness_vgg}
\end{table}
\begin{table*}[!htbp]
    \centering
    \begin{tabular}{ll|cccccccccccc}
    \toprule
        Model & XAI Method & Acc. & B.I. & CSDC & PC & DC & D & SD & TS & Com. & Cor. & Con & mX \\
        \hline
        {ViT-B/16} & OMENN (ours) & 0.98 & 1.00 & 0.89 & 0.85 & 0.85 & \textbf{0.92} & 0.74 & \textbf{0.97} & 0.89 & 0.74 & \textbf{0.97} & \textbf{0.87} \\
         & Chefer LRP & 0.98 & 1.00 & 0.91 & 0.92 & 0.89 & 0.90 & 0.74 & 0.95 & 0.90 & 0.74 & 0.95 & 0.86 \\
         & Rollout & 0.98 & 1.00 & 0.86 & 0.80 & 0.82 & 0.80 & 0.76 & 0.00 & 0.81 & 0.76 & 0.00 & 0.52 \\
         & IxG & 0.98 & 1.00 & 0.74 & 0.59 & 0.60 & 0.43 & 0.51 & 0.67 & 0.54 & 0.51 & 0.67 & 0.57 \\
         & IG & 0.98 & 1.00 & 0.89 & 0.86 & 0.85 & 0.90 & 0.65 & 0.91 & 0.88 & 0.65 & 0.91 & 0.82 \\
         & IG abs. & 0.98 & 1.00 & \textbf{0.96} & \textbf{0.98} & 0.95 & 0.89 & 0.63 & 0.74 & \textbf{0.92} & 0.63 & 0.74 & 0.76 \\
         & RISE & 0.98 & 1.00 & 0.79 & 0.71 & 0.70 & 0.83 & \textbf{0.79} & 0.75 & 0.78 & \textbf{0.79} & 0.75 & 0.77 \\
         & GradCAM & 0.98 & 1.00 & 0.75 & 0.67 & 0.68 & 0.91 & 0.70 & 0.48 & 0.81 & 0.70 & 0.48 & 0.66 \\
         & LIME & 0.98 & 1.00 & 0.95 & 0.96 &  \textbf{0.96} & 0.85 & 0.00 & 0.00 & 0.90 & 0.00 & 0.00 & 0.30 \\
         & B-cos & 0.96 & 0.87 & 0.93 & 0.88 & 0.94 & 0.86 & 0.69 & 0.89 & 0.89 & 0.69 & 0.89 & 0.82 \\
         \hline
         {VGG-16} & OMENN (ours) & 0.99 & 0.99 & 0.93 & 0.95 & 0.91 & 0.93 & 0.71 & \textbf{0.97} & 0.93 & 0.71 & \textbf{0.97} & \textbf{0.89} \\
         & IxG & 0.99 & 0.99 & 0.79 & 0.71 & 0.69 & 0.94 & 0.55 & 0.69 & 0.84 & 0.55 & 0.69 & 0.69 \\
         & IG & 0.99 & 0.99 & 0.92 & 0.92 & 0.92 & \textbf{0.97} & 0.67 & 0.92 & 0.92 & 0.67 & 0.92 & 0.85 \\
         & IG abs. & 0.99 & 0.99 & \textbf{0.96} & \textbf{0.99} & \textbf{0.97} & \textbf{0.97} & 0.69 & 0.84 & \textbf{0.97} & 0.69 & 0.84 & 0.83 \\
         & RISE & 0.99 & 0.99 & 0.80 & 0.73 & 0.70 & 0.84 & 0.73 & 0.83 & 0.79 & 0.73 & 0.83 & 0.78 \\
         & GradCAM & 0.99 & 0.99 & 0.94 & 0.97 & 0.93 & 0.87 & \textbf{0.75} & 0.93 & 0.91 & \textbf{0.75} & 0.93 & 0.86 \\
         & LIME & 0.99 & 0.99 & 0.89 & 0.88 & 0.90 & 0.92 & 0.00 & 0.00 & 0.91 & 0.00 & 0.00 & 0.30\\
         \hline
         {ResNet-50} & OMENN (ours) & 1.00 & 1.00 & 0.91 & 0.90 & 0.81 & 0.78 & 0.53 & 0.83 & 0.85 & 0.53 & 0.83 & 0.74 \\
         & IxG & 1.00 & 1.00 & 0.74 & 0.61 & 0.53 & 0.54 & 0.54 & 0.80 & 0.58 & 0.54 & 0.80 & 0.64 \\
         & IG & 1.00 & 1.00 & 0.92 & 0.94 & 0.88 &  \textbf{0.81} &  \textbf{0.59} &  \textbf{0.98} & 0.86 &  \textbf{0.59} &  \textbf{0.98} &  \textbf{0.81} \\
         & IG abs. & 1.00 & 1.00 & \textbf{0.95} &  \textbf{0.97} & 0.91 & 0.79 & 0.53 & 0.86 &  \textbf{0.87} & 0.53 & 0.86 & 0.75 \\
         & RISE & 1.00 & 1.00 & 0.82 & 0.75 & 0.74 & 0.63 & 0.56 & 0.61 & 0.70 & 0.56 & 0.61 & 0.62 \\
         & GradCAM & 1.00 & 1.00 & 0.80 & 0.74 & 0.69 & 0.74 & 0.55 & 0.78 & 0.74 & 0.55 & 0.78 & 0.69 \\
         & LIME & 1.00 & 1.00 & 0.94 & 0.94 &  \textbf{0.92} & 0.78 & 0.00 & 0.00 & 0.86 & 0.00 & 0.00 & 0.29\\
         \bottomrule
    \end{tabular}
    \caption{Results on FunnyBirds framework of different XAI methods with ViT-16/B, ResNet50 and VGG-16 backbones. OMENN achieves the best results among all methods for ViT and VGG, while for ResNet50 is slightly worse than IG abs and IG.}
    \label{tab:fb_results}
\end{table*}

\end{document}